\documentclass{article}
\usepackage{ijcai19}

\usepackage{times}
\usepackage{soul}
\usepackage{url}
\usepackage[utf8]{inputenc}
\usepackage[small]{caption}
\usepackage{graphicx}
\usepackage{booktabs}
\usepackage{xspace}
\usepackage[linesnumbered,vlined,ruled]{algorithm2e}
\usepackage{algpseudocode}
\SetKwInOut{KIN}{Input \ }
\SetKwInOut{KOUT}{Output \ }
\usepackage{amsmath, amsthm, amssymb}
\usepackage{array, multirow, booktabs}
\usepackage{subfigure}
\usepackage{hyperref}
\usepackage{supertabular}

\newcommand{\Solver}{\texttt{CoAPI}\xspace}
\newcommand{\Solverqx}{\texttt{CoAPI}-\texttt{qx}\xspace}
\newcommand{\Solverz}{\texttt{CoAPI}-\texttt{0it}\xspace}
\newcommand{\Solveri}{\texttt{CoAPI}-\texttt{1it}\xspace}
\newcommand{\Solverii}{\texttt{CoAPI}-\texttt{2it}\xspace}
\newcommand{\Solverzm}{\texttt{CoAPI}-\texttt{zm}\xspace}
\newcommand{\Primer}{\texttt{primer}\xspace}
\newcommand{\Primera}{\texttt{primer}-\texttt{a}\xspace}
\newcommand{\Primerb}{\texttt{primer}-\texttt{b}\xspace}

\newcommand{\CompileCover}{\textsc{CompileCover}\xspace}

\newcommand{\OverApproximate}{\textsc{OverApproximate}\xspace}

\newcommand{\Partition}{\textsc{Partition}\xspace}
\newcommand{\OrderSAT}{\textsc{OrderSAT}\xspace}

\newcommand{\Interval}{\textsc{Interval}\xspace}
\newcommand{\CompileAll}{\textsc{CompileAll}\xspace}

\newcommand{\cost}{\textsc{cost}\xspace}
\newcommand{\pri}{\textsc{prime}\xspace}

\newcommand{\True}{$\mathsf{true}$\xspace}

\newcommand{\UNSAT}{$\mathsf{UNSAT}$\xspace}

\newcommand{\ie}{{\it i.e.}}

\newcommand{\St}{{\it s.t.}}
\newcommand{\resp}{{\it resp.}}
\newcommand{\etc}{{\it etc.}}

\newcommand{\order}[1]{\langle #1 \rangle}

\newcommand{\AIP}{{AIP}}
\newcommand{\AC}{{AC}}
\newcommand{\PD}{{\pi}}
\newcommand{\F}{{\varphi}}
\newcommand{\CNF}{{\Sigma}}
\newcommand{\C}{{\mathcal{C}}}

\newcommand{\opie}{{\tau}}
\newcommand{\opc}{{\Omega}}
\newcommand{\DO}{{\Delta}}

\newcommand{\funFont}[1]{\textsc{#1}\xspace}

\title{\Solver: An Efficient Two-Phase Algorithm Using Core-Guided Over-Approximate Cover for Prime Compilation of Non-Clausal Formulae}

\author{
Weilin Luo$^1$
\and
Hai Wan$^1$\and
Hongzhen Zhong$^{1}$\And
Ou Wei$^2$
\affiliations
$^1$School of Data and Computer Science, Sun Yat-sen University\\
$^2$Department of Computer Science, University of Toronto
\emails
\{luowlin3, zhonghzh5\}@mail2.sysu.edu.cn,
wanhai@mail.sysu.edu.cn,
owei@cs.toronto.edu
}

\newtheorem{mydef}{Definition}
\newtheorem{thm}{Theorem}

\newtheorem{exam}{Example}

\begin{document}

\maketitle

\begin{abstract}
    \looseness = -1
    Prime compilation, \ie, the generation of all prime implicates or implicants ({\em primes} for short) of formulae, is a prominent fundamental issue for AI.
    Recently, the prime compilation for {\em non-clausal} formulae has received great attention.
    The state-of-the-art approaches proposed by~\cite{previti15} generate all primes along with a {\em prime} cover constructed by prime implicates using dual rail encoding. 
    However, the dual rail encoding potentially expands search space.
    In addition, constructing a prime cover, which is necessary for their methods, is time-consuming.
    To address these issues, we propose a novel {\em two-phase} method -- \Solver.
    The two phases are the key to construct a cover without using dual rail encoding.
    Specifically, given a non-clausal formula, we first propose a core-guided method to rewrite the non-clausal formula into a cover constructed by {\em over-approximate} implicates in the first phase. 
    Then, we generate all the primes based on the cover in the second phase.
    In order to reduce the size of the cover, we provide a {\em multi-order} based shrinking method, with a good tradeoff between the small size and efficiency, to compress the size of cover considerably.
    The experimental results show that \Solver outperforms state-of-the-art approaches.
    Particularly, for generating all prime implicates, \Solver consumes about one order of magnitude less time.
\end{abstract}

\section{Introduction}\label{sec:intr}
\looseness = -1
{\em Prime compilation} is a prominent fundamental issue for AI.
Given a non-clausal Boolean formula, prime compilation aims to generate all the primes of the formula. 
A prime does not contain redundant literals so that it can represent refined information.
Because of that, this problem has widely applications, including logic minimization~\cite{ignatiev15}, multi-agent systems~\cite{slavkovik14}, fault tree analysis~\cite{luo17}, model checking~\cite{bradley07}, bioinformatics~\cite{acuna12}, \etc

\looseness = -1
This problem is computationally hard.
For a non-clausal Boolean formula, the number of primes may be exponential in the size of the formula, while finding one prime is hard for the second level of the PH.
In practice, most problems can be hardly expressed in clausal formulae~\cite{stuckey13}.
Hence, non-clausal formulae are often transformed into CNF by some encoding methods, such as \emph{Tseitin encoding}~\cite{tseitin68}, which reduce the complexity by adding auxiliary variables.
Most of the earlier works only generate all primes of a CNF, but they cannot directly compute all primes of a non-clausal formula.
Therefore, this issue for non-clausal formula has received great attention~\cite{previti15}.

\looseness = -1
The state-of-the-art approaches~\cite{previti15} are capable of generating all primes of a non-clausal formula through several iterations.
They use \emph{dual rail encoding}~\cite{bryant87,roorda05} to encode search space in all produce.
Either a prime implicate or a prime implicant is computed at each iteration until a {\em prime} cover and all primes are obtained.
The cover is logically equivalent to the non-clausal formula, which guarantees that all the primes can be obtained.
Particularly, they extract a prime from an assignment based on the asymptotically optimal QuickXplain algorithm~\cite{junker04,bradley07}.

\looseness = -1
There are three issues in their methods. 
(i) The dual rail encoding with twice the number of variables than original encoding results in larger search space.
(ii) It is a time-consuming task to construct a prime cover because it should completely remove all redundant literals. 
(iii) It requires a minimal or maximal assignment in order to ensure the correctness, which often exerts a negative influence on SAT solving.
These issues probably explain that their performance on the inherent intractability of computing cases is still not satisfactory.
Notably, it is questionable whether finding a prime cover has a practical value since the influence of the size of the cover on the other parts of the algorithm is only vaguely known although the prime cover can be smaller.

\looseness = -1
We propose a novel two-phase method -- \Solver that focuses on an over-approximate cover with a good tradeoff between the small size of the cover and efficiency to improve the performance.
We stay within the idea of the work~\cite{previti15} that generates all primes based on a cover.
However, we use two separate phases to avoid using dual rail encoding in all phases.
We construct a cover without dual rail encoding in the first phase.
In the second phase, we generate all primes with that.
Furthermore, we introduce the notion of the {\em over-approximate implicate ({\AIP} for short)} that is an implicate containing as few literals as possible.
We consider constructing a cover with a set of {\AIP}s -- {\em over-approximate cover ({\AC} for short)}, rather than with a prime cover.
Note that an {\AC} is also logically equivalent to the non-clausal formula.

\looseness = -1
There are two challenges in our work.
The first one is the efficient computation of {\AIP}.
Motivated by the applications of the {\em unsatisfiable core} in optimizing large-scale search problems~\cite{narodytska14,yamada16}, we propose a core-guided method to produce {\AIP}s.
A smaller unsatisfiable core containing fewer literals should be efficiently obtained, which helps to reduce the number of {\AIP}s in the cover.
It is the second challenge, \ie, producing smaller unsatisfiable cores. 
We notice that the SAT solvers based on the two literal watching scheme~\cite{moskewicz01} cannot produce the smallest unsatisfiable core because of the limitation of the partial watching.
As for this, we provide a {\em multi-order} based shrinking method, in which we defined different decision orders to guide the shrinking of unsatisfiable cores in an iterative framework.
\looseness = -1
We evaluate \Solver on four benchmarks introduced by~\citeauthor{previti15}.
The experimental results show that \Solver exhibits better performance than state-of-the-art methods.
Especially for generating all prime implicates, \Solver is faster about one order of magnitude.

\looseness = -1
The paper is organized as follows.
Section~\ref{sec:prel} first introduces the basic concepts. 
Then, Section~\ref{sec:meth} and Section~\ref{sec:meth-shrink} present the main features of \Solver in detail.
After that, Section~\ref{sec:expe} reports the experiments.
Finally, Section~\ref{sec:rela} discusses related works and Section~\ref{sec:conc} concludes this paper.

\looseness = -1
Due to space limit, omitted proofs and supporting materials are provided in the additional file and online appendix ({\scriptsize\url{http://tinyurl.com/IJCAI19-233}}).


\section{Preliminaries}\label{sec:prel}

\looseness = -1
This section introduces the notations and backgrounds.


\looseness = -1
A {\em term} $\PD$ is a conjunction of literals, represented as a set of literals.
$|\PD|$ is the size of $\PD$, \ie, the number of literals in $\PD$.
Given a Boolean formula $\F$, a model is an assignment satisfying $\F$.
Particularly, a model is said to be {\em minimal} ({\resp} {\em maximal}), when it contains the minimal ({\resp} maximal) number of variables assigned \True.
A {\em clause} is the disjunction of literals, which is also represented by the set of its literals.
The size of a clause is the number of literals in it.
A Boolean formula is in conjunctive normal form (CNF) if it is formed as a conjunction of clauses, which denotes a set of clauses. For a Boolean formula $\CNF_{\F}$ in CNF, $|\CNF_{\F}|$ means the sum of the size of clauses in $\CNF_{\F}$.
Two Boolean formulae are {\em logically equivalent} iff they are satisfied by the same models.

\begin{mydef}
    \looseness = -1
    A clause $I_e$ is called an {\em implicate} of $\F$ if $\F \models I_e$. 
    Especially, $I_e$ is called {\em prime} if any clause $I_e'$ {\St} $I_e' \models I_e$ is not an implicate of $\F$.
\end{mydef}

\begin{mydef}
    \looseness = -1
    A term $I_n$ is called an {\em implicant} of $\F$ if $I_n \models \F$. 
    Especially, $I_n$ is called {\em prime} if any term $I_n'$ {\St} $I_n \models I_n'$ is not an implicant of $\F$.
\end{mydef}

\looseness = -1
The prime compilation aims to compute all the prime implicates or implicants, respectively, denoted by $PI^a_e$ and $PI^a_n$.


Given a Boolean formula $\F$, if $\F$ is unsatisfiable, a SAT solver based on CDCL, such as MiniSAT~\cite{een03a}, can produce a {\em proof of unsatisfiability}~\cite{mcmillan03a,zhang03a} using the resolution rule.

\begin{mydef}
    \looseness = -1
    A {\em proof of unsatisfiability} $\Pi$ for a set of clauses $\CNF_{\F}$ is a directed acyclic graph $(V_{\Pi}, E_{\Pi})$, where $V_{\Pi}$ is a set of clauses.
    For every vertex $c \in V_{\Pi}$, if $c \in \CNF_{\F}$, then $c$ is a {\em root}; 
    otherwise $c$ has exactly two predecessors, $c_l$ and $c_r$, such that $c$ is the resolvent of $c_l$ and $c_r$. 
    The empty clause, denoted by $\square$, is the unique {\em leaf}. 
\end{mydef}

\begin{mydef}
    \looseness = -1
    Given a proof of unsatisfiability $\Pi = (V_{\Pi}, E_{\Pi})$, for every clause $c \in V_{\Pi}$, the {\em fan-in cone} of $c$ includes of all the $c' \in V_{\Pi}$ from which there is at least one path to $c$.
\end{mydef}

\looseness = -1
A proof of unsatisfiability can answer what clauses are in the transitive fan-in cone of the empty clause. 
Therefore, an unsatisfiable core can be generated through backward traversing from $\square$.

\begin{mydef}
    \looseness = -1
    Given a Boolean formula $\CNF_{\F}$ in CNF, an {\em unsatisfiable core} $\C$ is a subset of $\CNF_{\F}$ and that is inconsistent.
\end{mydef}

\looseness = -1
The SAT solver, such as MiniSAT, is capable of handling assumptions. 
When the solver derives unsatisfiability based on the assumptions for a Boolean formula, it can return {\em failed assumptions}, which is a subset of assumptions inconsistent with the formula.
From here on, we use the terms {\em failed assumptions} and {\em unsatisfiable core} interchangeably since every unsatisfiable core corresponds to definite failed assumptions.
\section{New Approach -- \Solver}\label{sec:meth}
\looseness = -1
In this section, we first introduce the overview of \Solver, then show the details. 

\subsection{Overview}\label{sec:meth-over}
\looseness = -1
Given a Boolean formula $\F$, we first, based on original encoding, construct a cover in CNF to rewrite $\F$ in the first phase, \ie, the cover is logically equivalent to $\F$; then generate all primes in the second phase based on dual rail encoding.
Note that the two-phase produce not only avoids using dual rail encoding in all phases but also exploits the powerful heuristic branching method for SAT solving.
For simplicity, this paper only introduces the generation of all prime implicants of $\F$ (similarly for prime implicates because of the duality).

\looseness = -1
We extend the concepts of the prime implicate and prime cover into the {\AIP} and {\AC}, respectively, which are essential concepts in our algorithm and are defined as follows.

\begin{mydef}
    \looseness = -1
    An {\em over-approximate implicate} is a clause $\opie$ {\St} $\F \models \opie$.
    Given two over-approximate implicates $\alpha$ and $\beta$ of $\F$, if $\alpha \models \beta$, then we call $\alpha$ is {\em smaller} than $\beta$. 
\end{mydef}

\looseness = -1
The concept of {\AIP} is different from the concept of implicate because the former is as small as possible.
Notably, the prime implicate is a minimal {\AIP}.

\begin{mydef}
    \looseness = -1
    An {\em over-approximate cover} $\opc$ of $\F$ is a conjunction of over-approximate implicates of $\F$ and $\opc$ is logically equivalent to $\F$.
    The {\em cost} of an over-approximate cover $\opc$, denoted by \cost$(\opc) = |\opc|/|$\pri$(\F)|$.
\end{mydef}

\looseness = -1
Intuitively, \cost$(\opc)$ measures the degree of approximation of $\opc$ to \pri$(\F)$\footnote{\pri$(\F)$ returns a prime cover of $\F$.}. 

\begin{figure}[t]
    \centering
    \includegraphics[width=0.47\textwidth]{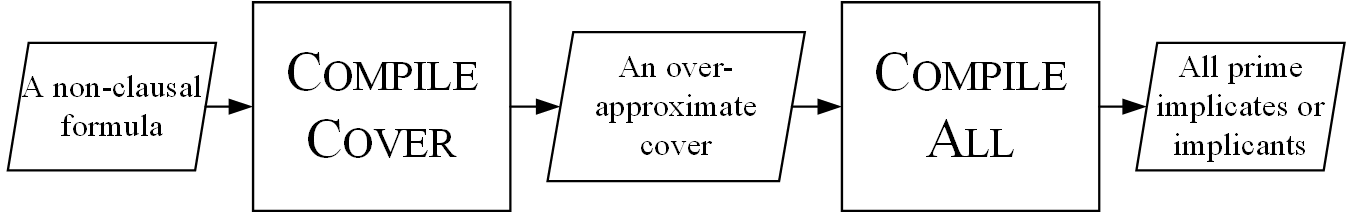}
    \caption{The framework of \Solver.}\label{fig:InPoPI_framework}
\end{figure}

\looseness = -1
The framework of \Solver includes two phases, namely \CompileCover and \CompileAll, which is shown in Figure~\ref{fig:InPoPI_framework}.
It takes a non-clausal Boolean formula $\F$ and its negation $\lnot \F$ as inputs.
The inputs are encoded as a set of clauses by Tseitin encoding or other methods.
Its output is all primes of $\F$.
\CompileCover first produces an {\AC} of $\F$ and then \CompileAll computes all primes.
We introduce the two phases in detail as follows.

\subsection{Core-Guided Over-Approximate Cover}\label{sec:meth-cover}

\looseness = -1
In order to construct a cover, the work~\cite{previti15} produces several prime implicates based on the QuickXplain algorithm.
A naive approach to extract a prime from an implicate, namely linear approach, is to linearly query whether it is still an implicate after flipping each literal of the implicate.
Therefore, the QuickXplain algorithm, based on recursively splitting the implicate, requires exponentially fewer queries in the optimal case than the linear approach.
However, it is still time-consuming for producing a prime implicate because there are considerable SAT queries to guarantee the prime.

\looseness = -1
In addition, the influence of the size of the cover on the other phases is only vaguely known.
Hence, although more computation time can lead to a smaller cover, it is not clear whether it is cost-effective in the overall algorithm.
The results of the Experiment~\ref{sec:expe_solver} demonstrate this view.
Based on the above considerations, we propose a core-guided method to produce {\AIP}s to rewrite $\F$.
It is possible to trade off the quality of the cover with the run time for extraction.

\begin{algorithm}[h]
    \caption{\funFont{CompileCover}}\label{alg:ccover}
    \KIN{A formula $\CNF_{\F}$ and its negation $\CNF_{\lnot \F}$ in CNF}
    \KOUT{An {\AC} $PI^c_e$}
    
    $\mathcal{R} \gets \CNF_{\lnot \F}$; $PI^c_e \gets \emptyset$\\
    
    \While{\True}
    {
        $(st$, $\PD) \gets $ \Call{SAT}{$\mathcal{R}$}\\
        \If{$st$ is \UNSAT}
        {
            \Return{$PI^c_e$}\\
        }
        \Else
        {
            $\PD_p \gets$ \Call{OverApproximate}{$\CNF_{\F}$, $\PD$}\\
            $PI^c_e \gets PI^c_e \cup \{\lnot \PD_p\}$\\
            $\mathcal{R} \gets \mathcal{R} \cup \{\lnot \PD_p\}$
        }
    }
    
\end{algorithm}

\looseness = -1
We construct a cover to rewrite $\F$ by iteratively computing {\AIP}s in \CompileCover shown in Algorithm~\ref{alg:ccover}.
To this end, \CompileCover maintains a set of clauses $\mathcal{R} = \CNF_{\lnot \F} \cup PI^c_e$, where $\CNF_{\lnot \F}$ encodes $\lnot \F$ in CNF ($\CNF_{\F}$ for $\F$) and $PI^c_e$ blocks already computed models.

\looseness = -1
We illustrate each iteration as follows.
\CompileCover first searches for a model $\PD$ of $\lnot \F$ which is not blocked by $PI^c_e$ (Line 3).
Then, \OverApproximate is invoked to shrink the unsatisfiable core of $\PD$ and $\F$ (Line 7). 
The more detail will be introduced in Section~\ref{sec:meth-shrink}.
After shrinking, \CompileCover updates $PI^c_e$ by adding $\lnot \PD_p$ (Line 8).
Clearly, $\lnot \PD_p$ is a smaller {\AIP} of $\F$ than $\lnot \PD$, since $\F \models \lnot \PD_p$ and $\lnot \PD_p \models \lnot \PD$.
In the end, the updated $\mathcal{R}$ prunes the search space for the next iteration (Line 9).

\looseness = -1
During the iterations, on the one hand, \CompileCover applies an incremental SAT solver to continually shrink the search space by conflict clauses. 
On the other hand, it also uses $PI^c_e$ to block the space that has been found.
Eventually, $PI^c_e$ prunes all the search space of $\lnot \F$, \ie, an {\AC} of $\F$ has been constructed by $PI^c_e$. 
At this point, $\mathcal{R}$ is unsatisfiable and the algorithm terminates.
We summarize an example for Algorithm~\ref{alg:ccover} as follows.


\begin{exam}
\looseness = -1
Given a formula $\F = (a \land b) \lor (\lnot a \land c)$, in the first iteration, a model $\lnot a \land b \land \lnot c$ of $\lnot \F$ is found; then, by consecutive SAT queries, we get a core $\lnot a \land \lnot c$; finally, an {\AIP} $a \lor c$ is produced, clearly, $\F \models a \lor c$.
During the same step, we can obtain a new {\AIP} $\lnot a \lor b$.
In total, $\mathcal{R}$ is unsatisfiable, where \CompileCover produces an {\AC} $(a \lor c) \land (\lnot a \lor b)$.
\end{exam}

\looseness = -1
For this example, \citeauthor{previti15} constructs the same result as us.
However, \Solver needs fewer SAT queries while their methods need more queries according to the size of implicate.
In general, \Solver reduces the number of SAT queries to speed up each iteration although it may take more iterations.

\subsection{Generation of All Primes}\label{sec:meth-all}

\looseness = -1
In \CompileAll, we encode $PI^c_e$ by dual rail encoding to initialize $\mathcal{H}$. 
Then, based on SAT solving, we iteratively compute all the minimal models of $\mathcal{H}$, \ie, all the prime implicants of $\F$.
This process is similar to~\cite{jabbour14}.
The more details about \CompileAll show in the additional file.

\section{Multi-Order based Shrinking}\label{sec:meth-shrink}


\looseness = -1
Constructing an {\AC} of $\F$ can be carried out iteratively to produce unsatisfiable cores.
Unfortunately, the SAT solver based on deterministic branching strategy often produces similar unsatisfiable cores for similar assumptions. 
In the worst case, the unsatisfiable core is the same size as the assumptions.
Therefore, it is worthwhile finding smaller unsatisfiable cores to compress the size of {\AC}.

\looseness = -1
Given a proof of unsatisfiability $\Pi$, an unsatisfiable core can be produced by traversing $\Pi$ backward.
Therefore, the generation of $\Pi$ determines the size of the unsatisfiable core.
We notice that the SAT solver based on the two literal watching scheme, which is powerful for SAT solving, selectively ignores some information during generating $\Pi$. 
We call this case {\em blocker ignoring} defined as follows.

\begin{mydef}
    \looseness = -1
    Given a Boolean formula $\CNF_{\F}$ in CNF and a proof of unsatisfiability $\Pi = (V_{\Pi}, E_{\Pi})$, the clause $\beta \in V_{\Pi}$ is a {\em blocker}, if $|\beta| = 1$ and $\beta$ is not a root. 
    An SAT solver {\em ignore} the satisfiability of the clauses containing $\beta$.
\end{mydef}

\begin{thm}\label{thm:blocker}
    \looseness = -1
    If a clause $\beta$ is a blocker of $\Pi = (V_{\Pi}, E_{\Pi})$, then there does not exist a clauses $c \in V_{\Pi}$ {\St} $\beta \models c$ unless $c$ is in the fan-in cone of $\beta$.
\end{thm}


\looseness = -1
Intuitively, if a blocker $\beta$ is generated, \ie, the literal $\beta$ is satisfied, then all clauses containing $\beta$ are naturally satisfied that can be ignored until backtracking to result in the freedom of $\beta$ in SAT solving. 
Therefore, these clauses do not appear in the $\Pi$ except the fan-in cone of $\beta$. 
This is powerful to search for a model because only the satisfiability of the necessary clauses needs to be considered, which is the core of the two literal watching scheme.
However, the blocker ignoring can miss important information for producing a small unsatisfiable core.
We use an example to explain this point.

\begin{figure}[t]
    \centering
    \includegraphics[width=0.47\textwidth]{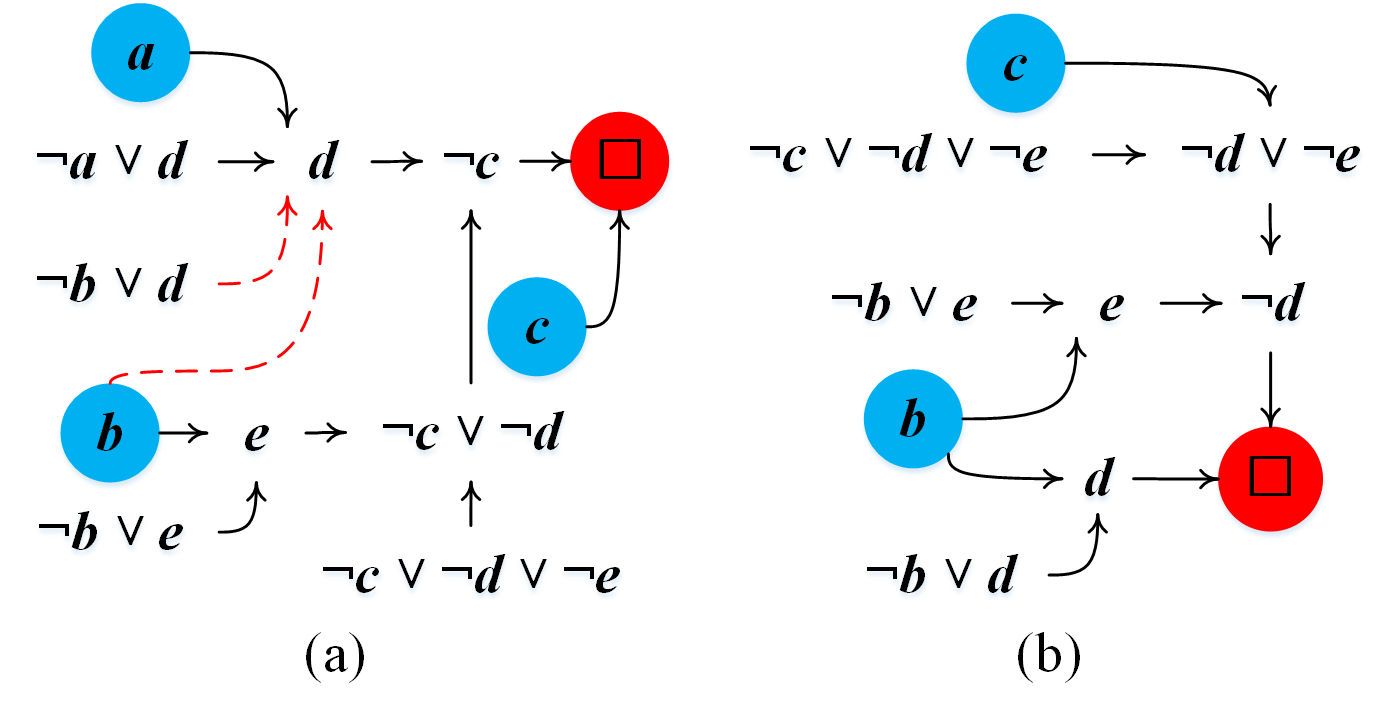}
    \caption{An example for the proof of unsatisfiability.}\label{fig:example_muti_order_shrink}
\end{figure}

\begin{exam}\label{exam:example_muti_order_shrink}
    \looseness = -1
    Given $\CNF_{\F} = (\lnot a \lor d) \land (\lnot b \lor d) \land (\lnot b \lor e) \land (\lnot c \lor \lnot d \lor \lnot e)$, the proof of unsatisfiability is shown in Figure~\ref{fig:example_muti_order_shrink}(a) based on the assumptions $a \land b \land c$.
    We can notice that the $\lnot b \lor d$ is missing due to blocker $d$, in which $d$ is not the resolvent of $b$ and $\lnot b \lor d$, but $a$ and $\lnot a \lor d$.
    This proof can produce the unsatisfiable core $a \land b \land c$.
\end{exam}

\looseness = -1
Disturbing decision order iteratively in the SAT solving is a useful and straightforward method to guide the smaller unsatisfiable core.
A similar approach was proposed by ~\cite{zhang03b}.
They iteratively invoke a SAT solver based on a random decision strategy to shrink an unsatisfiable core.
However, it lacks power for the prime compilation, which can be shown by the results of the Experiment~\ref{sec:expe_over}.
We propose a multi-order decision strategy instead of the random to iteratively shrink a small unsatisfiable core.
The multi-order decision strategy includes three kinds of decision orders defined as follows.

\begin{mydef}
    \looseness = -1
    A {\em decision order} is a list of variables $\DO = \order{...,x^i,...,x^j,...}$, in which $i < j$ and $x^i$ is picked earlier than $x^j$ by a SAT solver.
\end{mydef}

\begin{mydef}
    \looseness = -1
    Given an original decision order $\DO$, the {\em forward} decision order $\DO_f$ is the same as $\DO$.
    The {\em interval} decision order has two parts $\DO_l$ and $\DO_r$ with the following properties: 
    (i) if $x^i$ is in $\DO_l$ ({\resp} $\DO_r$), then $x^{i+1}$ is in $\DO_r$ ({\resp} $\DO_l$);
    (ii) $\forall x^i,x^j$ in $\DO_l$ ({\resp} $\DO_r$), if $x^i,x^j$ in $\DO$ {\St} $i < j$, then $x^i$ is also picked earlier than $x^j$ in $\DO_l$ ({\resp} $\DO_r$).
    The {\em backward} decision order $\DO_b$ is a reverse of $\DO$.
\end{mydef}

\looseness = -1
Our method allows a SAT solver to have the opportunity to produce a smaller unsatisfiable core from different definite orders.
Given a Boolean formula $\CNF_{\F}$ in CNF and its three variables $\alpha$, $\beta$, and $\gamma$, assume that a SAT solver can produce an unsatisfiable core of $\CNF_{\F}$ based on $\order{\alpha,\beta}$ while $\order{\alpha,\gamma}$ or $\order{\gamma,\alpha}$ can result in blockers to enlarge the size of the core.
Intuitively, for an original decision order $\order{\alpha,\beta,\gamma}$ ({\resp} $\order{\gamma,\alpha,\beta}$), based on the forward ({\resp backward}) decision order, \Solver can reduce the impact of blockers.
For an original order $\order{\alpha,\gamma,\beta}$, the impact can be lessened based on the interval decision decision order ($\DO_l = \order{\alpha,\beta}$, $\DO_r = \order{\gamma}$).

\begin{exam}
\looseness = -1
    Following the above Example~\ref{exam:example_muti_order_shrink}, the proof of unsatisfiability in Figure~\ref{fig:example_muti_order_shrink}(b) is based on $\DO_b$.
    In this case, a smaller unsatisfiable core, $a \land b$, can be produced.
\end{exam}

\looseness = -1
We provide a multi-order based shrinking method shown in Algorithm~\ref{alg:ovap}, in which \OrderSAT invokes a SAT solver with certain decision order.
The whole algorithm consists of two phases: {\em basic} and {\em iterative}.
In the basic phase, we first apply $\DO_f$ (Line 2), and then use $\DO_l$ and $\DO_r$ (Line 4).
In the iterative phase, we use $\DO_f$ (Line 7) and $\DO_b$ (Line 9) alternately until the bound of iterations or the fixpoint has been reached.
The fixpoint is that the size of the core does not change.

\begin{algorithm}[h]
    \caption{\funFont{OverApproximate}}\label{alg:ovap}
    \KIN{A CNF $\CNF_{\F}$ and a model $\PD$ of $\lnot \F$}
    \KOUT{An unsatisfiable core $\PD_p$}
    
    Initialize orders $\DO_f$, $\DO_l$, $\DO_r$, and $\DO_b$ based on the order of the index of variables in $\PD$\\
    $(st$, $\PD_p) \gets $ \Call{OrderSAT}{$\CNF_{\F}$, $\PD$, $\DO_f$}\\
    Update $\DO_l$ and $\DO_r$ based on $\PD_p$\\
    $\PD_p \gets $ \Call{Interval}{$\CNF_{\F}$, $\PD_p$, $\DO_l$, $\DO_r$}\\

    \While{The bound of iterations or the fixpoint has not been reached} {
        \If{the last order is $\DO_b$} {
            $(st$, $\PD_p) \gets $ \Call{OrderSAT}{$\CNF_{\F}$, $\PD_p$, $\DO_f$}\\
        }\Else{
            $(st$, $\PD_p) \gets $ \Call{OrderSAT}{$\CNF_{\F}$, $\PD_p$, $\DO_b$}\\
        }
    }

    \Return{$\PD_p$}\\
\end{algorithm}

\begin{algorithm}[h]
    \caption{\funFont{Interval}}\label{alg:inte}
    \KIN{A CNF $\CNF_{\F}$, a model $\PD$ of $\lnot \F$, $\DO_l$, and $\DO_r$.}
    \KOUT{An unsatisfiable core $\PD_p$}
    
    $\PD_p \gets \PD$\\
    \If{$|\PD_p|$ is $1$}{
        \Return{$\PD_p$}\\
    }
    \Else{
        $(\PD_l$, $\PD_r) \gets$ \Call{Partition}{$\PD_p$, $\DO_l$, $\DO_r$}\\
        $(st$, $\C) \gets$ \Call{OrderSAT}{$\CNF_{\F}$, $\PD_l$, $\DO_l$}\\
        \If{$st$ is \UNSAT}{
            Update $\DO_l$ and $\DO_r$ based on $\C$\\
            \Return{\Call{Interval}{$\CNF_{\F}$, $\C$, $\DO_l$, $\DO_r$}}\\
        }
        \Else{
            $(st$, $\C) \gets$ \Call{OrderSAT}{$\CNF_{\F}$, $\PD_r$, $\DO_r$}\\
            \If{$st$ is \UNSAT}{
                Update $\DO_l$ and $\DO_r$ based on $\C$\\
                \Return{\Call{Interval}{$\CNF_{\F}$, $\C$, $\DO_l$, $\DO_r$}}\\
            }
            \Else{
                \Return{$\PD_p$}\\
            }
        }
    }
    
\end{algorithm}



\looseness = -1
Based on the interval decision order, we partition the unsatisfiable core to explore better results.
Algorithm~\ref{alg:inte} summarizes \Interval that is similar to the QuickXplain algorithm, in which \Partition partitions an unsatisfiable core based on $\DO_l$ and $\DO_r$.
Compared with the QuickXplain algorithm, \Interval avoids discussing the case where none of $\PD_l$ and $\PD_r$ is a model of $\lnot \F$ to cut down the time consumption (Line 16).

Note that \OrderSAT with $\DO_l$ or $\DO_r$ is potentially harder than that with $\DO_f$ or $\DO_b$. 
The reasons are as follows.
First, \OrderSAT with the assumptions $\PD$ in \Interval, in which $\PD \models \lnot \F$ is unknown, is in NPC.
Second, based on $\DO_f$ or $\DO_b$, \OrderSAT with the assumptions $\PD$, in which $\PD \models \lnot \F$ holds, is in polynomial time.
Hence, we only use the interval decision order in the basic phase while apply $\DO_f$ and $\DO_b$ in the two phases.


\section{Experimental Results}\label{sec:expe}

\begin{figure*}[t]
    \centering
    \subfigure[Computation of prime implicate.]{\includegraphics[width=0.33\textwidth]{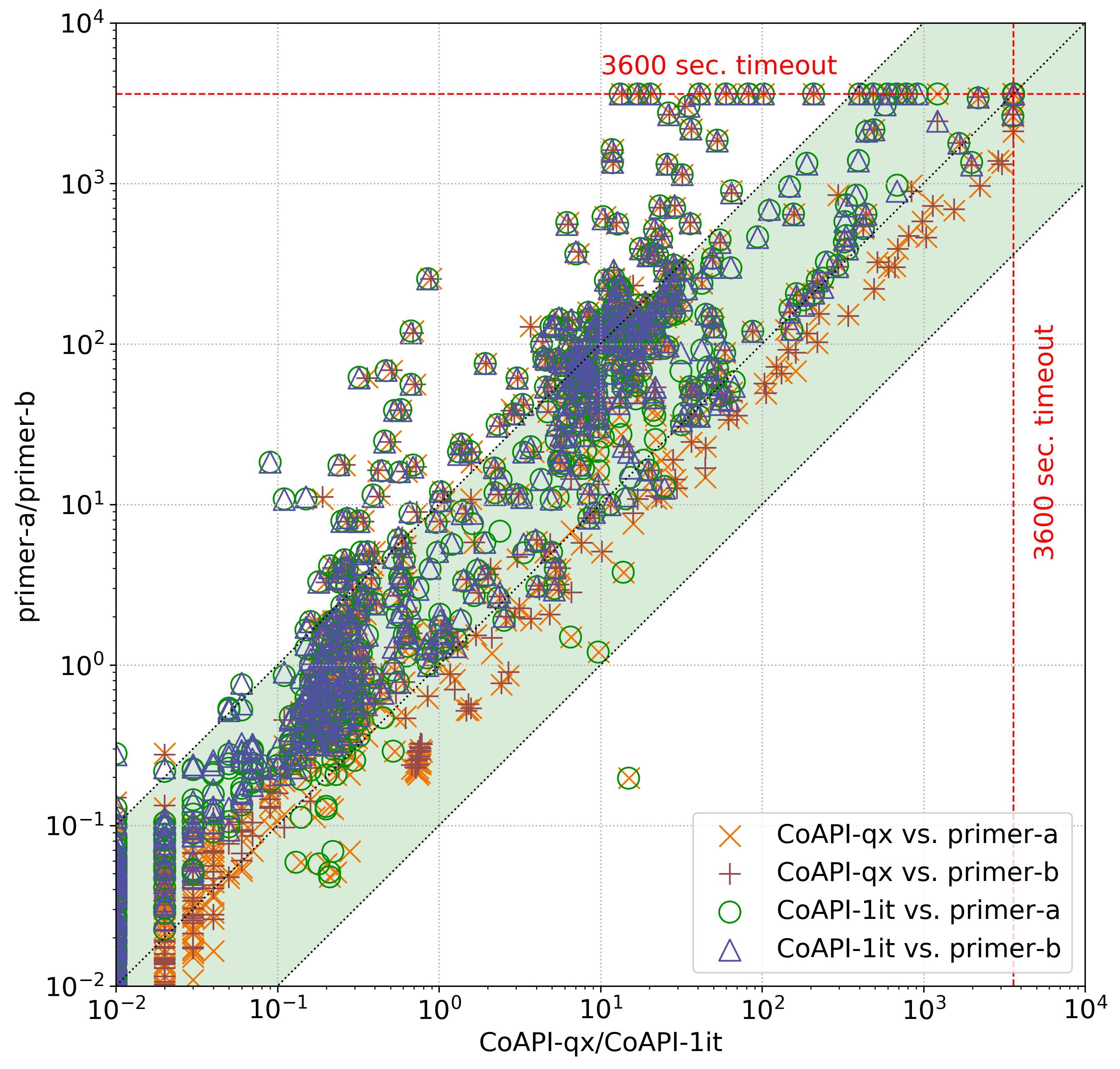}}
    \subfigure[Computation of prime implicant.]{\includegraphics[width=0.33\textwidth]{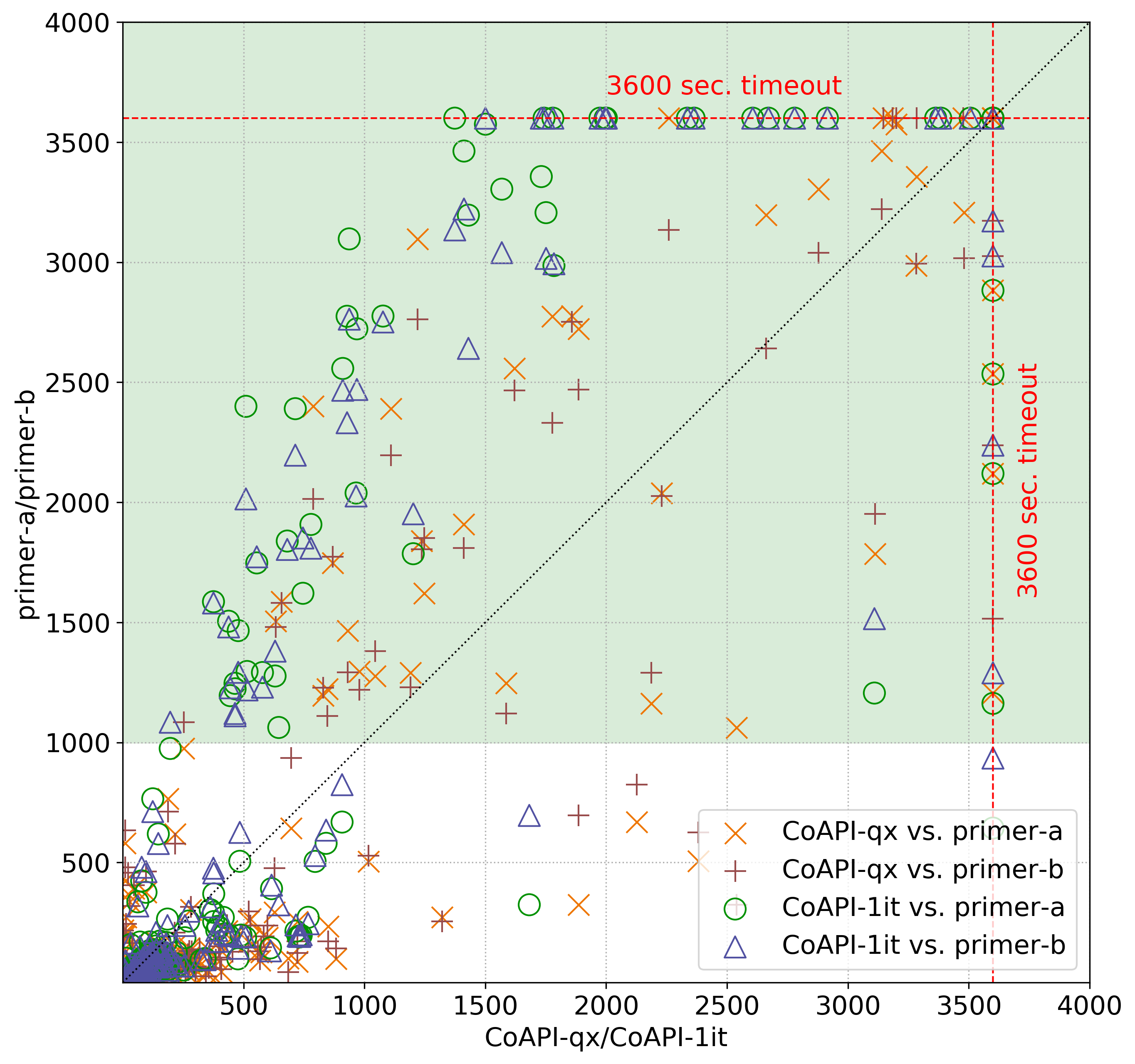}}
    \subfigure[Shrinking results.]{\includegraphics[width=0.33\textwidth]{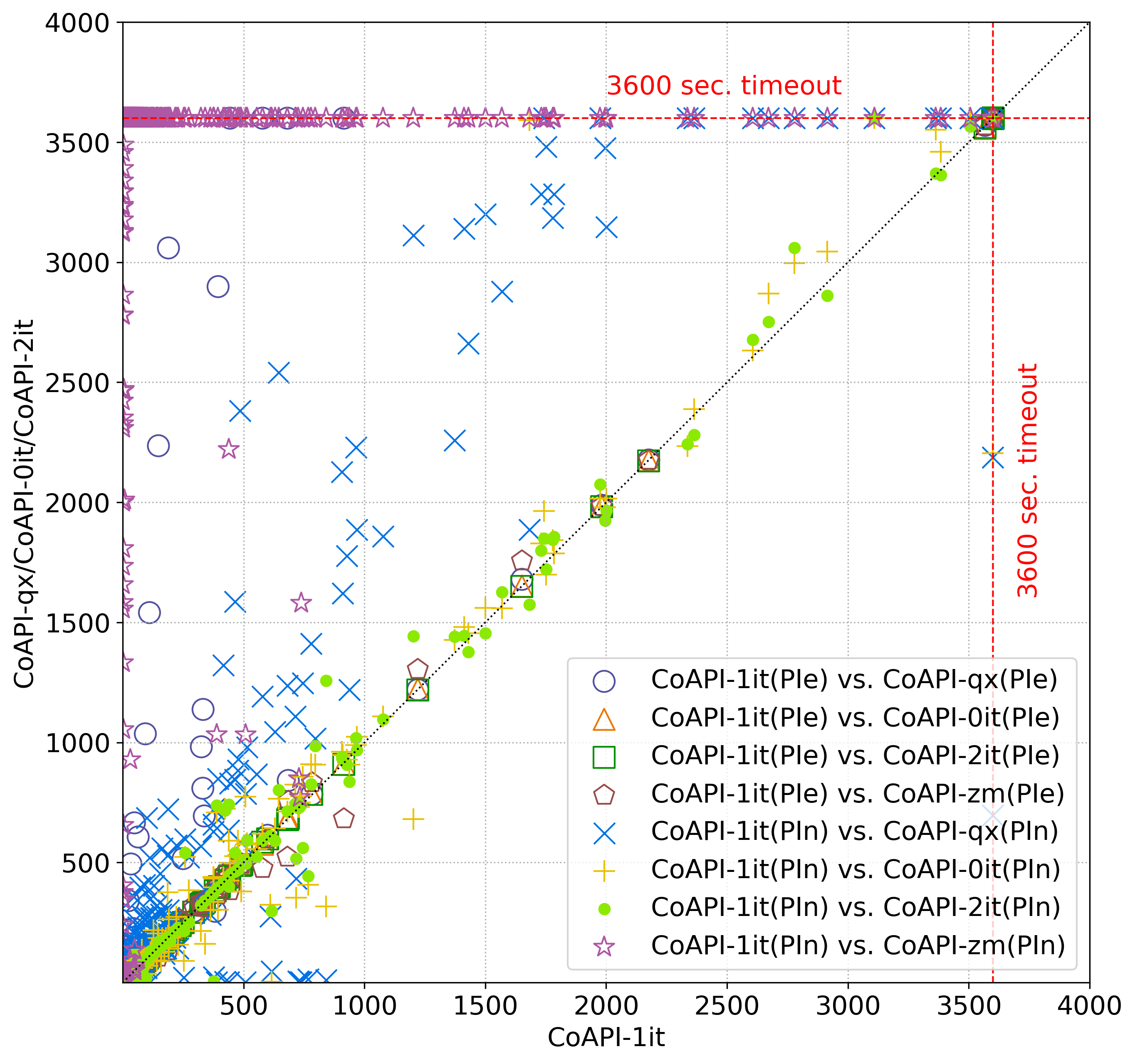}}
    \caption{Performance comparison.}\label{fig:exam}
    
\end{figure*}

\looseness = -1
To evaluate our method, we compared \Solver and its variants with the state-of-the-art methods -- \Primera and \Primerb~\cite{previti15}\footnote{https://reason.di.fc.ul.pt/wiki/doku.php?id=-primer.} over four benchmarks, and discussed the effects of different shrinking strategies.
In each experiment, we considered two tasks: (i) generating all prime implicates; (ii) generating all prime implicants.
We implemented \Solver utilizing MiniSAT\footnote{https://github.com/niklasso/minisat.} that was also used to implement \Primera and \Primerb.
The benchmarks are introduced by~\citeauthor{previti15}, denoted by \emph{QG6}, \emph{Geffe gen.}, \emph{F+PHP}, and \emph{F+GT}, respectively.
The experiments were performed on an Intel Core i5-7400 3 GHz, with 8 GByte of memory and running Ubuntu. 
For each case, the time limit was set to 3600 seconds and the memory limit to 7 GByte.


\subsection{Comparisons between \Solver and \Primer}\label{sec:expe_solver}

\looseness = -1
We assess the performance of \Solver in this section.
In this experiment, we implemented the variant of \Solver, denoted by \Solverqx, which uses the QuickXplain algorithm to construct a prime cover in the first phase.
We also implemented \Solver with only one iteration, denoted by \Solveri.
We evaluate the performance of \Solverqx, \Solveri, \Primera, and \Primerb by the 743 cases.

\begin{table}[h]
    \center
    \caption{The number of solved cases.}\label{tab:num_solved_pi}
    \scriptsize

    \begin{tabular}{|*{6}{c|}}
        \hline
            &\emph{QG6} &\emph{Geffe gen.}  &\emph{F+PHP}   &\emph{F+GT}    &\emph{Total} \\
            &(83)       &(600)              &(30)           &(30)           &(743) \\
        \hline
        \hline
            \Primera    &30 / 66           &576 / \textbf{596} &30 / 30  &28 / 30           &664 / 722\\
        \hline
            \Primerb    &30 / 65           &577 / \textbf{596} &30 / 30  &28 / 30           &665 / 721\\
        \hline
        \hline
            \Solverqx   &30 / 70           &\textbf{589} / 592 &30 / 30  &26 / 30           &675 / 722\\
        \hline
            \Solveri    &30 / \textbf{81}  &\textbf{589} / 591 &30 / 30  &\textbf{30} / 30  &\textbf{679} / \textbf{732}\\
        \hline
    \end{tabular}
\end{table}

\looseness = -1
Table~\ref{tab:num_solved_pi} shows the number of cases that can be computed.
The results are separated by the symbol `/', on the right of which is for the task (i) and the left of which is for the task (ii). 
It is used for all tables.
Overall, \Solverqx and \Solveri can successfully solve more cases than \Primera and \Primerb.
Note that the 679 cases solved by \Solveri include all the 664 ({\resp} 665) ones solved by \Primera ({\resp} \Primerb) in the task (i).
It is obvious that, for \emph{QG6}, \Solverqx and \Solveri dramatically increase the number of cases successfully solved in the task (ii).

\looseness = -1
The more detailed comparisons of these methods for the task (i) are shown in Figure~\ref{fig:exam}(a).
The X-axis indicates the time in seconds taken by \Solverqx or \Solveri, and the Y-axis indicates that taken by \Primera or \Primerb.
Points above the diagonal indicate advantages for \Solverqx or \Solveri.
\Solveri generally computes much faster than \Primera ({\resp} \Primerb) in 92\% ({\resp} 96\%) cases -- it consumes about at least one order of magnitude less time than \Primera ({\resp} \Primerb) in 26\% ({\resp} 27\%) cases.
For \Solverqx, the advantage is still obvious.
It is in 73\% ({\resp} 80\%) cases that \Solverqx beats \Primera ({\resp} \Primerb), in which \Solverqx is about one order of magnitude faster in 18\% ({\resp} 19\%) cases than \Primera ({\resp} \Primerb).
In this task, most of the literals in implicate are necessary.
Therefore, the QuickXplain algorithm may require significantly more SAT queries than our method.

\looseness = -1
Figure~\ref{fig:exam}(b) shows the performance of these methods for the task (ii) in detail.
Cases that are negative for \Solveri focus on \emph{F+PHP} and \emph{F+GT}, because the prime covers of these formulae are in the form $(x_1 \lor y_1) \land ... \land (x_m \lor y_m)$ that is extremely beneficial to generate all primes.
We focus on the challenging cases that are computed over 1000s by \Primera or \Primerb, \ie, the cases are shown in the green area in Figure~\ref{fig:exam}(b). 
Most of the points above the diagonal (at least 62\% cases for \Solverqx and 84\% for \Solveri) indicate the advantage of our methods.
In particular, \Solveri dominates \Primera and \Primerb on \emph{QG6} reducing used time for at least 40.26\%.

\begin{table}[h]
    \center
    \caption{The improvement of our methods for challenging cases.}\label{tab:improve}
    \scriptsize
    
    \begin{tabular}{|*{3}{c|}}
        \hline
            &Win    &Win x10+\\
        \hline
        \hline
            \Solverqx vs. \Primera  &70\% / 64\%                       &47\% / 0\%\\
        \hline
            \Solverqx vs. \Primerb  &70\% / 62\%                       &47\% / 0\%\\
        \hline
            \Solveri vs. \Primera &\textbf{93\%} / \textbf{86\%}     &\textbf{50\%} / 0\%\\
        \hline
            \Solveri vs. \Primerb &\textbf{93\%} / \textbf{84\%}     &\textbf{50\%} / 0\%\\
        \hline
    \end{tabular}
\end{table}

\looseness = -1
For challenging cases, the improvements of our methods are shown in Table~\ref{tab:improve}, in which the columns present the percentage of faster cases (Win) and the percentage of at least one order of magnitude faster cases (Win x10+).
Note that, for \Solveri, the number of faster cases in the task (ii) increases to 86\% ({\resp} 84\%) and the number of cases with at least one order of magnitude faster improves to 50\% ({\resp} 50\%) in the task (i).

\looseness = -1
In general, our methods outperform the state-of-the-art methods, particularly in the task (i).
The outstanding performance of \Solverqx shows that the two-phases framework is efficient because it avoids using dual rail encoding and the minimal or maximal assignment strategy throughout the whole algorithm.
Moreover, we can notice that \Solveri is better than \Solverqx because of the {\AC}, which is described in the next section.

\subsection{Evaluations of Over-Approximation}\label{sec:expe_over}

\looseness = -1
To evaluate the different shrinking strategies, we implemented \Solverz without iterations and \Solverii with two iterations. 
Moreover, \Solverzm uses the strategy proposed by ~\cite{zhang03b}. 
Based on our experiences, \Solverzm with 11 iterations gives the best performance for the two tasks in practice. 

\begin{table}[h]
    \center
    \caption{Results of shrinking unsatisfiable cores.}\label{tab:shrink}
    \scriptsize
    
    \begin{tabular}{|*{5}{c|}}
        \hline
            &Cost   &Fixpoint   &First Shrink &Other Shrink\\
        \hline
        \hline
            \Solverz    &1.00 / 2.59    &$-$ / $-$      &7\% / 92\% &7\% / 93\%\\
        \hline
            \Solveri    &1.00 / 1.75    &0\% / 0\%      &7\% / 92\% &7\% / 94\%\\
        \hline
            \Solverii   &1.00 / 1.72    &99\% / 74\%    &7\% / 92\% &7\% / 94\%\\
        \hline
        \hline
            \Solverzm   &1.00 / 6854.80  &100\% / 99\%   &7\% / 65\% &7\% / 67\%\\
        \hline
    \end{tabular}
\end{table}

\looseness = -1
We compare \Solver and \Solverzm in different shrinking strategies on the same benchmarks as above.
The results are shown in Figure~\ref{fig:exam}(c).
The most points are above the diagonal line, which represents a less used time for \Solveri in most cases.
\Solverii and \Solverzm are comparable in the task (i).
However, in the task (ii), \Solverzm only solves 302 of 743 cases that are all simple for \Solveri.

    

\looseness = -1
Table~\ref{tab:shrink} shows the statistics on average for shrinking unsatisfiable cores.
Due to \Solverqx with a prime cover, we compute the cost of {\AC}s based on the \Solverqx.
The columns present the cost (Cost), the ratio of reaching the fixpoint (Fixpoint), the ratio of the shrinking size in the first time (First Shrink), and the ratio of the shrinking size in the other times (Other Shrink).

\looseness = -1
The generally lower costs of \Solverz, \Solveri, and \Solverii show the shrunk unsatisfiable core can be much smaller in all cases.
From the statistics, the cost is often reduced by running the shrinking procedure iteratively, but usually, the gains for the shrinking core are not as substantial as the first shrinking.
This point is also reflected in the ratio of shrinking size, in which the size of the {\AIP}s reduces dramatically in the first time, but not by much during the following shrinkings.
We also note that \Solverii can reach a fixpoint in most cases.
These statistics mean that the only one iteration is the best tradeoff for the quality of the unsatisfiable core with the run time for these benchmarks.
Comparing \Solverii with \Solverzm, the cost and the ratio of the shrinking size in the task (ii) illustrate that \Solverzm cannot effectively control shrinking, while \Solverii does well for it.




\section{Related Works and Discussions}\label{sec:rela}

\looseness = -1
Many of techniques to the prime compilation are based on branch and bound/backtrack search procedures~\cite{castell96,ravi04,dharbe13,jabbour14}. 
They take full advantage of powerful SAT solvers, while these methods cannot generate the primes for non-clausal formulae.
In addition, a number of approaches based on {\em binary decision diagrams (BDD)}~\cite{coudert92} or {\em zero-suppressed BDD (ZBDD)}~\cite{simon01} have been proposed.
These methods can encode primes in a compact space thanks to BDD. 
Given the complexity of the problem, however, these methods may still suffer from time or memory limitations in practice.
Almost simultaneously, a {\em 0-1 integer linear programming (ILP)} formulation~\cite{pizzuti96,manquinho97,silva97,palopoli99} was proposed to compute primes of CNF formulae.
Although these approaches can naturally encode the minimal constraints utilizing ILP, their efficiency is questionable.

\looseness = -1
Most present works~\cite{castell96,pizzuti96,manquinho97,silva97,palopoli99,ravi04,jabbour14} only focus on computing primes of CNF or DNF, while there are also some approaches for working on non-clausal formulae.
Such as, \citeauthor{nagir93} \shortcite{nagir93} studied a more general algorithm for prime implicate generation, which allows any conjunction of DNF formulae. The approaches based on the BDD/ZBDD can compute prime implicants of non-clausal formulae.
Additionally, \citeauthor{ramesh97} \shortcite{ramesh97} computed prime implicants and prime implicates of NNF formula.
Recently, \citeauthor{previti15} \shortcite{previti15} described the most efficient approach at present.
\looseness = -1
In order to produce a small {\AIP}, we need to generate small unsatisfiable cores.
\citeauthor{zhang03b} \shortcite{zhang03b} produced small unsatisfiable cores by the random order and multiple iterations.
This approach is similar to our idea, but they fail to find the relationship between the order and the size of unsatisfiable cores, resulting in their approach without the ability to further shrink unsatisfiable cores.
\citeauthor{gershman06} \shortcite{gershman06} suggested a more effective shrinking procedure based on dominators resulting from the proof of unsatisfiability.
Naturally, analysis based on proof of unsatisfiability increases the cost of a single iteration.
Hence, considering large-scale iterations for shrinking different unsatisfiable cores in our work, their method does not work well to this.

\section{Conclusions and Future Works}\label{sec:conc}

\looseness = -1
We have proposed a novel approach -- \Solver for the prime compilation based on unsatisfiable cores.
Compared with the work~\cite{previti15}, \Solver separates the generating processes into two phases, which can permit us to construct a cover without using dual rail encoding resulting in shrinking the search space.
Moreover, we have proposed a core-guided approach to construct an {\AC} to rewrite the formula.
It should emphasize that the {\AC} can be efficiently computed.
Besides, we have provided a multi-order based method to shrink a small unsatisfiable core.
The experimental results have shown that \Solver has a significant advantage for the generation of prime implicates and better performance for prime implicants than state-of-the-art methods.

\looseness = -1
For future work, we expect that our method can be applied to the task of producing a small proof of unsatisfiability.



\bibliographystyle{named}
\bibliography{paper}

\end{document}